\documentclass[conference,final]{IEEEtran}
\usepackage{times}

\usepackage{microtype}

\usepackage[numbers]{natbib}
\usepackage{multicol}

\usepackage[bookmarks=true,hidelinks]{hyperref}

\usepackage{amssymb}
\usepackage{amsmath}
\usepackage{mathtools}
\usepackage{physics}
\usepackage{siunitx}
\usepackage[acronym,shortcuts]{glossaries}
\glsdisablehyper

\usepackage{float}
\usepackage[above,below]{placeins}
\usepackage{graphicx}
\usepackage[x11names]{xcolor}
\usepackage{tikz}
\usetikzlibrary{3d}
\usetikzlibrary{calc}
\usetikzlibrary{math}
\usetikzlibrary{arrows.meta}
\usetikzlibrary{decorations}
\usetikzlibrary{decorations.pathmorphing}
\usetikzlibrary{decorations.pathreplacing}
\usetikzlibrary{positioning}
\usepgflibrary{fpu}
\usepackage{pgfplots}
\pgfplotsset{compat=1.17}

\usepackage[ruled,longend,linesnumbered]{algorithm2e}
\SetArgSty{textnormal}

\SetCommentSty{mycommfont}

\usepackage[obeyFinal]{easy-todo}
\usepackage{adjustbox}

\usepackage{ifdraft}
\ifoptionfinal{}{
    \usepackage[switch]{lineno}
    \linenumbers
}

\usepackage[nameinlink,noabbrev,capitalize]{cleveref}

\newacronym{cuda}{CUDA}{Compute Unified Device Architecture}
\newacronym{simt}{SIMT}{Single-Instruction, Multiple-Thread}
\newacronym{gpu}{GPU}{Graphics Processing Unit}
\newacronym{cas}{CAS}{Atomic Compare-and-Swap}
\newacronym{dem}{DEM}{Discrete Element Method}
\newacronym{mpm}{MPM}{Material Point Method}
\newacronym{ncp}{NCP}{Nonlinear Complementarity Problem}
\newacronym{pgs}{PGS}{Projected Gauss-Seidel}
\newacronym{pja}{PJA}{Projected Jacobi Algorithm}
\newacronym{sdf}{SDF}{Signed Distance Function}
\newacronym{dof}{DoF}{Degree of Freedom}
\newacronym{lammps}{LAMMPS}{Large-scale Atomic/Molecular Massively Parallel Simulator}
\newacronym{liggghts}{LIGGGHTS}{LAMMPS Improved for General Granular and Granular Heat Transfer Simulations}
\newacronym{td3}{TD3}{Twin Delayed Deep Deterministic Policy Gradient}

\newcommand{\R}{\mathbb{R}}
\newcommand{\SEThree}{SE\left(3\right)}
\newcommand{\seThree}{se\left(3\right)}
\newcommand{\RigidTransform}[2]{{}_{#1}X^{#2}}
\newcommand{\InCoordinates}[2]{{}_{#1}{#2}}
\newcommand{\NParticles}{n_p}
\newcommand{\NContacts}{n_c}
\newcommand{\NBodies}{n_b}
\newcommand{\Pos}{x}
\newcommand{\Vel}{v}
\newcommand{\Impulse}{\Delta \Vel}
\newcommand{\DT}{\Delta t}

\newcommand{\texttask}[1]{\textsc{#1}}
\newcommand{\textpardefn}[1]{\emph{#1}}

\newcommand{\bOne}{b\texttt{[1]}}
\newcommand{\bTwoThr}{b\texttt{[2:3]}}

\colorlet{MyRed}{LightSalmon2}
\colorlet{MyPink}{PeachPuff1}
\colorlet{MyBlue}{LightSkyBlue1}
\colorlet{MyYellow}{Goldenrod1!50!Ivory1}
\colorlet{MyGreen}{DarkOliveGreen2}
\colorlet{MyGrey}{Snow3}
\colorlet{MyOffWhite}{Ivory1}
\def\spacing{0.05}

\tikzstyle{global}=[]
\tikzstyle{linethread}=[
global,
decorate,
decoration={snake,segment length=1.5mm, amplitude=0.3mm},
MyGrey
]
\tikzstyle{arrowtime}=[global,linethread,thin,black,-{Kite[fill=white,length=3mm]}]
\tikzstyle{boxsubroutine}=[global,rounded corners=0.75mm]
\tikzstyle{textsubroutine}=[global,node font=\scriptsize]
\tikzstyle{lineloopbrace}=[global,decorate,decoration={brace,amplitude=2mm}]
\tikzstyle{textloopbrace}=[global,midway,above=2.5mm,node font=\small]
\tikzstyle{axisline} = [-Computer Modern Rightarrow,thick,line cap=butt]

\pgfplotsset{every axis/.append style={
            mark size=1.5pt,
            cycle multiindex* list={
                    MyRed!80!black,
                    MyBlue!80!black,
                    MyYellow!80!black,
                    MyGreen!80!black
                    \nextlist
                    mark list\nextlist
                    linestyles*
                },
            minor tick num=3,
            tick style={black},
            legend style={nodes={scale=0.8}},
            legend cell align={left},
            grid style={black!10},
            axis background style={fill={white}},
        }}

\pgfplotsset{every axis plot/.append style={
            enlargelimits={rel=0.08},
            thick,
            line cap=butt,
            line join=miter,
        }}

\pgfplotsset{speedup plot/.style={
            ymajorgrids=true,
            ytick distance=1,
            yticklabel={
                    \pgfmathprintnumber[fixed relative, precision=0]{\tick}x
                }
        }}

\pgfplotsset{x axis kilo/.style={
            xtick=data,
            xticklabel={
                    \pgfkeys{/pgf/fpu=true}
                    \pgfmathparse{int(exp(\tick)/1000)}%
                    \pgfmathfloatifflags{\pgfmathresult}{0}{
                        \pgfmathparse{int(ceil(exp(\tick)))}
                        \pgfmathprintnumber[fixed relative, precision=0]{\pgfmathresult}
                    }{
                        \pgfmathprintnumber[fixed relative, precision=0]{\pgfmathresult}k
                    }
                    \pgfkeys{/pgf/fpu=false}
                },
        }}

\pgfplotsset{
    discard if not/.style 2 args={
            x filter/.code={
                    \edef\tempa{\thisrow{#1}}
                    \edef\tempb{#2}
                    \ifx\tempa\tempb
                    \else
                        \def\pgfmathresult{inf}
                    \fi
                }
        }}

\begin{document}

\title{GranularGym: High Performance Simulation for Robotic Tasks with Granular Materials}

\author{\authorblockN{David Millard\authorrefmark{1},
        Daniel Pastor\authorrefmark{2},
        Joseph Bowkett\authorrefmark{2},
        Paul Backes\authorrefmark{2} and
        Gaurav S.\ Sukhatme\authorrefmark{1}}
    \authorblockA{\authorrefmark{1}University of Southern California, Los
        Angeles, USA. Email: \texttt{dmillard@usc.edu}}
    \authorblockA{\authorrefmark{2}Jet Propulsion Lab, Pasadena, USA}}

\maketitle
\global\csname @topnum\endcsname 0
\global\csname @botnum\endcsname 0
\ifoptionfinal{}{
    \thispagestyle{plain}
    \pagestyle{plain}
}

\begin{abstract}
    Granular materials are of critical interest to many robotic
tasks in planetary science, construction, and manufacturing. However, the
dynamics of granular materials are complex and often computationally very
expensive to simulate. We propose a set of methodologies and a system for the
fast simulation of granular materials on \glspl{gpu}, and show that this
simulation is fast enough for basic training with Reinforcement Learning
algorithms, which currently require many dynamics samples to achieve acceptable
performance. Our method models granular material dynamics using implicit
timestepping methods for multibody rigid contacts, as well as algorithmic
techniques for efficient parallel collision detection between pairs of particles
and between particle and arbitrarily shaped rigid bodies, and programming
techniques for minimizing warp divergence on \gls{simt} chip architectures. We
showcase our simulation system on several environments targeted toward robotic
tasks, and release our simulator as an open-source tool.
\end{abstract}

\IEEEpeerreviewmaketitle

\section{Introduction}
\label{sec:introduction}
\begin{figure}[t]
    \centering
    \begin{tikzpicture}
        \clip [draw,boxsubroutine] (0, 0) rectangle coordinate (centerpoint) (\columnwidth,7.5cm);
        \node [inner sep=0pt] at (centerpoint) {\includegraphics[width=\columnwidth]{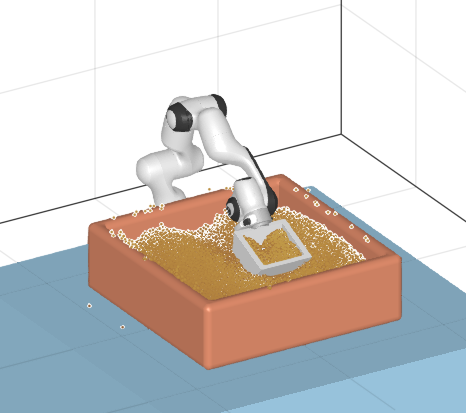}};
        \draw [boxsubroutine] (0, 0) rectangle (\columnwidth,7.5cm);
    \end{tikzpicture}
    \caption{
        Simulation of a Franka Emika Panda robot with a rigidly attached scoop
        attachment, interacting with a bed of 50000 particles, running at
        realtime on a single NVIDIA GeForce RTX 3080 Ti with a simulated
        timestep of \num{5e-4}. Our simulation engine approximates granular
        material state and dynamics as a system of rigidly interacting spherical
        particles, which may interact with kinematically driven rigid bodies of
        arbitrary geometry.
    }
    \label{fig:hero-image}
\end{figure}

Robots are expanding their operational domains from structured environments,
like factory floors, into the unstructured world of
homes~\cite{yamamotoDevelopmentHumanSupport2019}, the
outdoors~\cite{fernandezInformativePathPlanning2022}, and other planets in our
solar system~\cite{farleyMars2020Mission2020}. Successful operation in any environment requires a robot to predict the effect of its actions on the world,
so that it may select an action that best achieves its goal.

The dynamics of the objects in a robot's environment may be quite complex. In
addition to rigid objects, robots may encounter
deformable~\cite{antonovaBayesianTreatmentRealtosim2022} or
sliceable~\cite{heiden2021disect} objects including
cloth~\cite{salhotraLearningDeformableObject2022} or
ropes~\cite{limReal2Sim2RealSelfSupervisedLearning2022}. Here, our focus is on
robots interacting with large quantities of granular material like rock, sand,
or loose soil, and which are given tasks that require prediction of the bulk
state of such materials, such as material transport or shaping.

While machine learning is a powerful and generalizable tool for robot prediction
and perception~\cite{fuDeepWholeBodyControl2022}, it remains necessary to use
large models and a large number of samples to achieve accurate performance. Physical
modeling, which we refer to in this paper as simulation, is a well-tested and
interpretable method for dynamical system prediction.  Robust and efficient
simulation provides robotics engineers with accurate, safe, and fast
environments to test algorithms, train data-driven learning agents, and serve as
a predictive model in robotic control algorithms.

To these ends, we have developed \texttt{GranularGym}, a faster-than-realtime
simulation engine for the mechanics of granular materials with tens
of thousands of particles and at interactive speeds for hundreds of thousands
of particles, running on a single commodity \gls{gpu}. Additionally, we
contribute a set of environments that we utilize as benchmarks for algorithms
intending to solve granular material manipulation tasks. We
document the equations of motion for the simulated dynamics, describe the
algorithms and data structures used for high-performance simulation on
\glspl{gpu}, analyze the performance of our engine and how it scales with
various parameters, and present several benchmark environments and associated
scaling of performance, and show that our engine is performant enough to train
state-of-the-art reinforcement learning algorithms to achieve high rewards.

Our approach uses a simplified model of granular interaction based on rigid body
interparticle contact, which may not capture the full, vast space of rheological
phenomena found in nature. Nevertheless, we believe that fast, approximate
simulation in complex domains is a powerful tool for the development of robotic
autonomy, particularly in closed-loop robotic systems where modeling errors may
be accounted for and corrected based on sensor observations.

The code for \texttt{GranularGym} is released under an open-source license and
is written in Julia~\cite{bezanson2017julia}, with a
Python interface available, and targets multithreaded CPU and GPU
architectures that support the NVIDIA \gls{cuda} or Apple Metal frameworks, and
is easily portable to other parallel compute platforms.

\section{Methods}
\label{sec:methods}
We simulate rigid, dry-frictional contact of $\NParticles$ spherical particles
and $\NBodies$ rigid bodies of arbitrary geometry interacting in a scene. The
particle states and velocities are described by matrices $\Pos, \Vel \in
    \R^{3\times n}$. Each rigid body's state is given by a six \gls{dof}
transform $\RigidTransform{0}{i} \in \SEThree$ from the world frame $0$ into
the frame of body $i$. Rigid body $i$ also has velocity $v_i \in \seThree$,
where $\seThree$ is the Lie algebra of the Special Euclidean Lie group
$\SEThree$. Rigid bodies in our engine are fully driven by time and may exert
forces on granular particles but do not accumulate a reaction force.

In the following subsections, we describe our algorithm for computing contact
impulses across large systems of particles, efficient parallel data structures
for broadphase collision detection and non-convex rigid body collision
detection, and branched execution considerations for efficient programming in
\gls{simt} \gls{gpu} environments.

\subsection{Implicit Contact Impulses with Projected Jacobi}
\label{subsec:projected-jacobi}

We compute particle impulses with rigid dry frictional contact using an implicit
timestepping approach common in rigid body simulation by constructing a
\gls{ncp} to solve for the systemic contact impulse. A thorough treatment and
comparison of implicit timestepping methods is given in
\citet{horakSimilaritiesDifferencesContact2019a}. We use a parallelized variant
of the serialized \gls{pgs} method to solve this \gls{ncp}, which we call the
\gls{pja}, listed in \Cref{alg:projected-jacobi}.  For completeness, in this
section, we detail the computation of the contact impulse \gls{ncp} and then
describe the \gls{pja} solver and its implementation on a \gls{cuda} \gls{gpu}.

\begin{figure}[t]
    \centering
    \def\xtilt{1/4}
\def\ytilt{1/4}
\def\frameroll{-15}

\begin{tikzpicture}[z={(\xtilt,-\ytilt)}]

\def\contactradius{2}
\def\coneangle{65}
\def\coneheight{1.5}
\def\axislength{2.0}

\tikzstyle{frictioncone} = [fill=MyYellow]
\tikzstyle{spheredim} = [thin]
\tikzstyle{spheredimbehind} = [spheredim,dotted]
\tikzstyle{impulseline} = [semithick]
\tikzstyle{impulsefill} = [fill=black,fill opacity=0.1]

\begin{scope}[rotate=\frameroll]
\tikzmath{
\conewidth=\coneheight*tan(\coneangle/2);
\h=\coneheight;
\ry=\conewidth;
\rx=\xtilt*\conewidth;
coordinate \a, \b;
\a = (-{\h+\rx*(\rx/\h)},-{\ry*sqrt(1-(\rx/\h)*(\rx/\h))});
\b = (-{\h+\rx*(\rx/\h)},{\ry*sqrt(1-(\rx/\h)*(\rx/\h))});
coordinate \redc, \bluec;
\redc = ({-\contactradius*cos(\frameroll)},{-\contactradius*sin(\frameroll)});
\bluec = ({\contactradius*cos(\frameroll)},{\contactradius*sin(\frameroll)});
coordinate \imp, \projimp;
\projimp=($0.7*(\a)$);
\imp=($(\projimp)+(0, -0.6)$);
{
\filldraw[fill=MyRed]
    (-\contactradius, 0)
    circle[radius=\contactradius];
\filldraw[fill=MyBlue]
    (\contactradius, 0)
    circle[radius=\contactradius];
\begin{scope}[shift only]
\draw[spheredimbehind]
    (\redc)+(\contactradius, 0) arc[
        x radius=\contactradius,
        y radius=\ytilt*\contactradius,
        start angle=0,
        end angle=180
    ];
\draw[spheredimbehind]
    (\bluec)+(\contactradius, 0) arc[
        x radius=\contactradius,
        y radius=\ytilt*\contactradius,
        start angle=0,
        end angle=180
    ];
\end{scope}

\filldraw[frictioncone,semithick] (0, 0) -- (\a) -- (\b) -- cycle;
\filldraw[frictioncone,thin]
    (-\coneheight, 0)
    circle[x radius=\rx, y radius=\ry];
\draw[semithick]
    (-\coneheight, -\ry)
    arc[x radius=\rx, y radius=\ry, start angle=-90, end angle=90];
\draw[impulseline] (0, 0) -- (\imp) node[at end, below left] {$\Impulse^*$};
\draw[impulseline] (0, 0) -- (\projimp);
\draw[impulseline,dashed,-] (\imp) -- (\projimp);
\path[impulsefill] (0, 0) -- (\imp) -- (\projimp) -- cycle;
\draw[axisline,-,draw opacity=0.2] (0, 0, 0) -- (-\coneheight+\rx, 0, 0);
\draw[axisline]
    (-\coneheight+\rx, 0, 0) -- (-\axislength, 0, 0)
    node[at end, left] {$e_1$};
\draw[axisline,shift only]
    (0, 0, 0) -- (0, 0, \axislength)
    node[at end, below right] {$e_2$};
\draw[axisline]
    (0, 0, 0) -- (0, \axislength, 0)
    node[at end,above] {$e_3$};
\begin{scope}[shift only]
\draw[spheredim]
    (\redc)+(\contactradius, 0) arc[
        x radius=\contactradius,
        y radius=\ytilt*\contactradius,
        start angle=0,
        end angle=-180
    ];
\draw[spheredim]
    (\bluec)+(\contactradius, 0) arc[
        x radius=\contactradius,
        y radius=\ytilt*\contactradius,
        start angle=0,
        end angle=-180
    ];
\end{scope}
};
}
\end{scope}

\end{tikzpicture}
    \caption{
        Graphical representation of a contact frame between two objects (red and
        blue).  The contact normal is aligned with the frame's $e_1$-axis, while
        the $e_2$ and $e_3$ axes are an arbitrary basis for the frictional force
        plane. A cone representing the bounds of dry Coulomb friction is shown
        in yellow.  $\Impulse^*$ shows a candidate impulse, and the dashed line
        shows its projection onto the friction cone via the solution of the
        contact \gls{ncp} described in \Cref{subsec:projected-jacobi}.
    }
    \label{fig:contact-frame}
\end{figure}
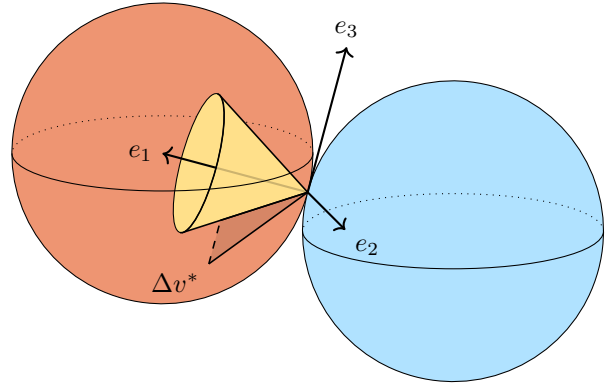

\begin{algorithm}[t]
    \caption{Projected Jacobi Algorithm}
    \label{alg:projected-jacobi}

    $\Impulse \gets 0$\;

    \ForEach{iteration from $1$ to $n$}{
        \ForEach{contact pair $(i, j, \psi)$}{
            $b
                \gets
                J_{i,j}^T (
                v^{(k)}_i
                - \gamma v^{(k)}_j
                + \Delta t F_{ext,i}
                + \Impulse_i
                )
            $\;
            $\bOne \gets \max(\bOne + \frac{\alpha \psi}{\Delta t}, 0)$\;
            \tcp{Project onto Coulomb friction cone.}
            \If{$||\bTwoThr||_2 > \mu \cdot (\bOne)$}{
                $\bTwoThr \gets \mu \cdot (\bOne) \frac{\bTwoThr}{||\bTwoThr||_2}$\;
            }
            $\Impulse_i \gets \Impulse_i + J_{i,j}b$\;
        }
    }
\end{algorithm}

During collision impulse calculation, it is useful to establish a \emph{contact
    frame}, shown in \Cref{fig:contact-frame}. A contact frame is a local
coordinate system with the origin at the point of contact and the $e_1$-axis
oriented along the contact normal. As a result, the contact normal force is
axis-aligned with $e_1$, and frictional forces exist only in the $e_2e_3$
plane.  For contact between particles $i$ and $j$, the \emph{contact
    Jacobian} $J_{i,j} \in R^{3 \times 3}$ is the transformation into the
associated contact frame. We assemble these into a large, sparse contact
Jacobian for the entire system, $J \in \R^{\NContacts \times \NParticles}$,
where $\NContacts$ is the number of contacts at the current timestep.  The
contact impulses $\Impulse$ is then computed from the solution to
\cref{eqn:contact-impulse}, where $M$ is the mass-inertia matrix, $v^{(k)}$ is
the system velocity at timestep $k$, $c^{(k)}$ is the post-contact relative
velocity in the contact frame at timestep $k$, $F_{ext}$ is the summary of any
external forces applied to the system, such as gravity, and $\Delta t$ is the
timestep.
\begin{equation}
    \label{eqn:contact-impulse}
    c^{(k+1)}
    =
    \underbrace{J M^{-1} J^T}_{A} \Impulse
    +
    \underbrace{J(v^{(k)} + M^{-1} \Delta t F_{ext})}_b
\end{equation}
To enforce the non-penetration of rigid body contact, where objects may either
exert a contact force on one another, or accelerate away from one another, but
not both, we add the complementarity constraint in
\cref{eqn:complementarity-constraint} to \cref{eqn:contact-impulse} on the
normal directions
\begin{equation}
    \label{eqn:complementarity-constraint}
    0 \le \Impulse \cdot e_1
    \quad \bot \quad
    (A \Impulse + b) \cdot e_1 \ge 0
\end{equation}
Furthermore, to enforce a Coulomb approximation of dry friction, we include the
nonlinear constraint in \cref{eqn:friction-constraint} between the tangent and
normal impulses
\begin{equation}
    \label{eqn:friction-constraint}
    ||\mathrm{proj}_{e_2e_3} \Impulse||_2 \le \mu \Impulse \cdot e_1,
\end{equation}
where $\mu$ is the coefficient of friction and $\mathrm{proj}_{e_2e_3}$ is a
projection operator into the $e_2e_3$ plane. The projection onto the friction
cone is the nonlinearity in the \gls{ncp}. Together,
\Cref{eqn:contact-impulse,eqn:complementarity-constraint,eqn:friction-constraint}
form the systemic contact impulse \gls{ncp}. The post separation contact
velocities $c^{(k)}$ are treated as slack variables, and we use the resulting
contact impulses $\Impulse$ in a symplectic Euler time integration scheme, where
the updated velocity is computed first, per \Cref{eqn:sym-euler-step1}, and then
used to update positions, per \Cref{eqn:sym-euler-step2}.
\begin{align}
    \Vel^{(k+1)} = \Vel^{(k)} + \DT\Impulse \label{eqn:sym-euler-step1} \\
    \Pos^{(k+1)} = \Pos^{(k)} + \Vel^{(k+1)} \label{eqn:sym-euler-step2}
\end{align}
We use a symplectic Euler integrator for its energy preserving properties,
particularly in systems of dynamics with rigid body
contacts\cite{erezSimulationToolsModelbased2015}.

Due to accumulating errors inherent in simulation, particularly in implicit
methods, particles may intersect by small amounts. Since the contact impulse
\gls{ncp} of \Cref{eqn:contact-impulse} is specified in terms of velocity and
impulse, rather than position, such errors would accumulate if left untreated.
To mitigate erroneous penetration, we apply a stabilization scheme proposed by
\citet{baumgarteStabilizationConstraintsIntegrals1972a}, by applying a small
corrective impulse along the contact normal of magnitude $\frac{\alpha
        \psi}{\Delta t}$, where $\psi$ is the penetration depth, and $\alpha$ is a user
chosen parameter.

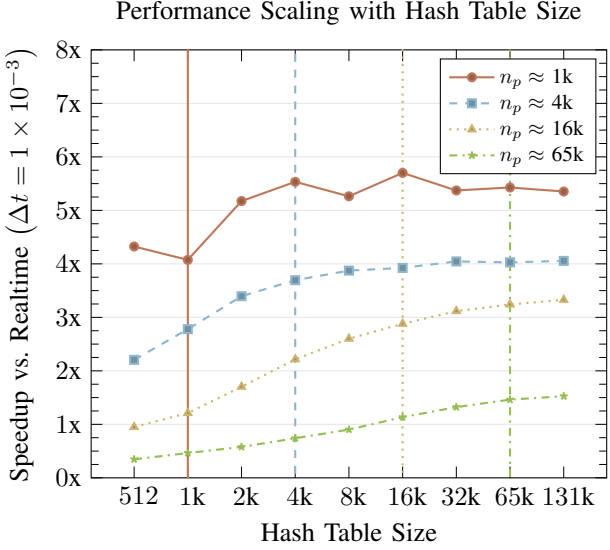
\begin{figure}[t]
    \centering
    \newcommand\plotnparticles[1]{
    \addplot table[
        col sep=comma,
        x={hash_table_size},
        y=rt_factor_np_#1,
    ] {figures/data/hashtable_archimedes_benchmark_NVIDIA_GeForce_RTX_3080_Ti.csv};
}

\begin{tikzpicture}
\begin{semilogxaxis}[
    speedup plot,
    x axis kilo,
    title style={align=center},
    title={Performance Scaling with Hash Table Size},
    xlabel={Hash Table Size},
    ylabel={Speedup vs. Realtime $\left(\Delta t = \num{1e-3}\right)$},
    ymin={0},
    ymax={8},
]
    \foreach \nparticles in {1024,4096,16384,65536} {
        \pgfkeys{/pgf/fpu=true,/pgf/fpu/output format=fixed}
        \tikzmath{ \npthoufp = int(\nparticles/1000); }
        \pgfkeys{/pgf/fpu=false}
        \tikzmath{ \npthou = int(\npthoufp); }
        
        \plot+[forget plot] coordinates {(\nparticles, -10)(\nparticles,20)};

        \edef\dolegend{\noexpand\addlegendentry{$\NParticles \approx$ \npthou k}} 
        \plotnparticles{\nparticles};
        \dolegend
    }
\end{semilogxaxis} 
\end{tikzpicture}
    \caption{
        Plot of speedup over realtime against hash table size (higher numbers
        are better). To maximize the number of collision candidates, we simulate
        particles at rest inside a tall cylindrical column. The vertical lines
        mark the number of particles used in the performance plot of
        the corresponding style.  Based on the shape of these performance curves, we
        propose a heuristic hash table size of twice the number of particles
        simulated.
    }
    \label{fig:hashtable-benchmark}
\end{figure}

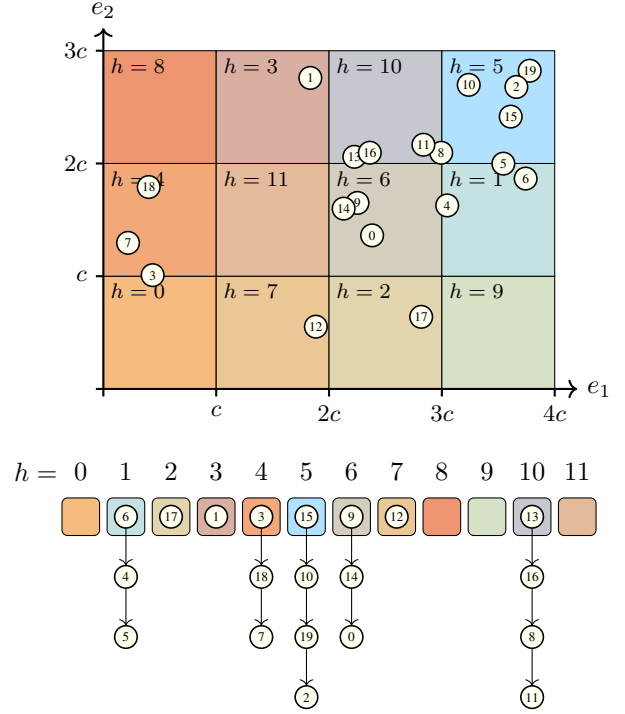
\begin{figure}[t]
    \centering
    \begin{tikzpicture}

\def\ticklen{0.1}
\def\ncellsw{4}
\def\ncellsh{3}
\def\nparticles{20}
\def\particleseed{3}
\def\hc{0.6}
\def\sc{1.5}
\def\ll{0.8}
\def\particleradius{0.15}
\def\topleftcolor{MyRed}
\def\toprightcolor{MyBlue}
\def\botleftcolor{MyYellow}
\def\particlecolor{MyOffWhite}

\tikzset{xycoloredbox/.style 2 args = {
    boxsubroutine,
    fill=\toprightcolor!#1!\topleftcolor!#2!\botleftcolor,
}}
\tikzstyle{hashcelllabel} = []
\tikzstyle{tickline} = [thick]
\tikzstyle{ticklabel} = [node font=\small, at end]
\tikzstyle{particle} = [fill=\particlecolor,semithick]
\tikzstyle{particlelabel} = [node font=\tiny]
\tikzstyle{particlearrow} = [-Computer Modern Rightarrow]

\tikzmath{
function hash(\x, \y) {
    \i = int(\ncellsw * \y + \x);
    \h = int(mod(103 * \i, \ncellsw * \ncellsh));
    return \h;
};
function reorder(\i) {
    \h = int(mod(103 * int(\i), \nparticles));
    return \h;
};
}

\tikzmath{
for \y in {1,...,\ncellsh}{
    for \x in {1,...,\ncellsw}{
        \xfrac=int(100 * (\x-1) / (\ncellsw-1));
        \yfrac=int(50 + 50 * (\y-1) / (\ncellsh-1));
        \h=hash(\x-1,\y-1);
        \htxfrac{\h}=\xfrac;
        \htyfrac{\h}=\yfrac;
        {
            \filldraw[xycoloredbox={\xfrac}{\yfrac},sharp corners]
                ({\sc*(\x-1)}, {\sc*(\y-1)})
                rectangle
                (\sc*\x, \sc*\y);
            \node[hashcelllabel,below right,node font=\footnotesize]
                at ({\sc*(\x-1)}, {\sc*\y})
                {$h=\h$};
        };
    };
};
{
    \draw[axisline]
        (-\ticklen, 0) -- (\sc*\ncellsw+3*\ticklen, 0)
        node[at end, right] {$e_1$};
};
for \x in {1,...,\ncellsw}{
    let \xstr = ;
    if \x > 1 then { \xstr = int(\x); };
    {\draw[tickline] (\sc*\x, 0) -- (\sc*\x, -\ticklen) node[ticklabel, below] {${\xstr}c$};};
};
{
    \draw[axisline]
        (0, -\ticklen) -- (0, \sc*\ncellsh+3*\ticklen)
        node[at end, above] {$e_2$};
};
for \y in {1,...,\ncellsh}{
    let \ystr = ;
    if \y > 1 then { \ystr = int(\y); };
    {\draw[tickline] (0, \sc*\y) -- (-\ticklen, \sc*\y) node[ticklabel, left] {${\ystr}c$};};
};
}

\tikzmath{
\n=\ncellsh*\ncellsw;
\htx=(\sc*\ncellsw-\hc*\n)/2;
\hty=-2.0;
{\node[hashcelllabel] at (\htx-\hc/2, \hty+3*\hc/2) {$h=$};};
for \h in {0,...,\n-1}{
    \hi=int(\h);
    \hcount\hi=0;
    \x=\hc+\htx+\hc*\h;
    \y=\hc+\hty;
    {
        \filldraw[xycoloredbox={\htxfrac\hi}{\htyfrac\hi}]
            (\x-\hc+\spacing, \y-\hc+\spacing)
            rectangle
            (\x-\spacing, \y-\spacing);
        \node[hashcelllabel] at (\x-\hc/2, \y+\hc/2) {$\hi$};
    };
};
}

\pgfmathsetseed{\particleseed};
\tikzmath{
for \i in {1,...,\nparticles}{
    \ii=reorder(\i);
    \x=\particleradius + random()*(\sc*\ncellsw - 2*\particleradius);
    \y=\particleradius + random()*(\sc*\ncellsh - 2*\particleradius);
    \xx=floor(\x/\sc);
    \yy=floor(\y/\sc);
    \h=hash(\xx,\yy);
    \lx=\hc/2+\htx+\hc*\h;
    \ly=\hc/2+\hty-\ll*\hcount\h;
    \hcount\h = int(\hcount\h + 1);
    {
        \filldraw[particle] (\x, \y) circle[radius=\particleradius];
        \node[particlelabel] at (\x, \y) {\ii};
        \filldraw[particle] (\lx, \ly) circle[radius=\particleradius];
        \node[particlelabel] at (\lx, \ly) {\ii};
    };
    if \hcount\h > 1 then {
        {\draw[particlearrow] (\lx, \ly+\ll-\particleradius) -- (\lx, \ly+\particleradius);};
    };
};
}

\end{tikzpicture}
    \caption{
        Graphical representation of the spatial hashmap data structure for fast,
        parallel broadphase collision checking. Physical coordinates are divided
        into cells of size $c$, each with a hash value $h$ computed from the
        cell's location. A linked hash table, shown in the lower half of the
        figure, is constructed in parallel according to
        \Cref{alg:parallel-hashmap-constructor}, with one thread for each
        particle in the scene.  This data structure generalizes to any number of
        dimensions; we show a two-dimensional representation for clarity.
    }
    \label{fig:spatial-hashmap-graphic}
\end{figure}

\begin{algorithm}[b]
    \caption{Parallel Spatial Hashmap Construction}
    \label{alg:parallel-hashmap-constructor}
    \SetKwFunction{AtomicCAS}{AtomicCAS}
    \SetKwFunction{AddressOf}{AddressOf}
    \SetKwFunction{SpatialHash}{SpatialHash}
    \SetKwFunction{Round}{Round}

    \ForEach{particle $i$}{
        $\bar{x}_i \gets$ \Round{$\frac{x_i}{2r}$}\;
        $h_i \gets$ \SpatialHash{$\bar{x}_i, n_h$}\;
        $p \gets$ \AddressOf{$table[h]$}\;
        $head \gets table[h_i]$\;
        \While{$head = table[h_i]$}{
            $head \gets$ \AtomicCAS{$p, head, i$ }\;
        }
        $next[i] \gets head$\;
    }
\end{algorithm}

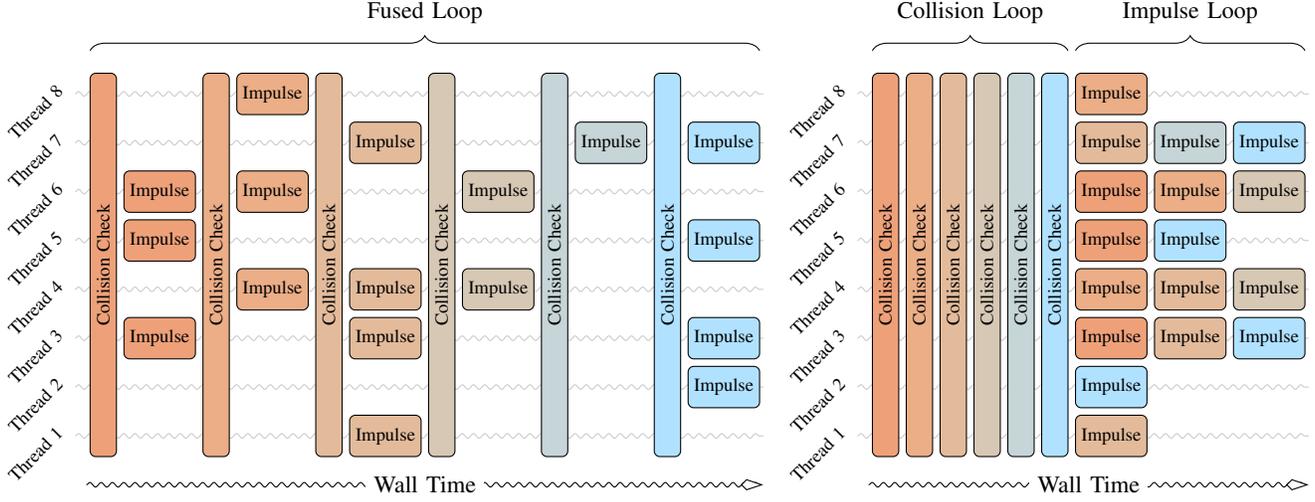
\begin{figure*}[t]
    \centering
    \begin{tikzpicture}

\def\nthreads{8}
\def\ntimesteps{6}
\def\wcol{0.45}
\ifx\wimp\undefined \def\wimp{1.05} \fi
\def\hthread{0.65}
\def\figseed{2}
\ifx\vertstack\undefined \def\vertstack{0} \fi
\def\braceoffset{0.25}
\def\tstartcol{MyRed}
\def\tmidcol{MyYellow}
\def\tendcol{MyBlue}

\tikzstyle{coloredboxsubroutine}=[boxsubroutine,fill=\tendcol!#1!\tmidcol!#1!\tstartcol]

\pgfdeclarelayer{bg}
\pgfsetlayers{bg,main}

\def\drawthreadlines{
    \begin{pgfonlayer}{bg}
    \foreach \thread in {1,...,\nthreads}{
        \draw[linethread]
            (-0.15,{\hthread*(\thread+0.5)}) -- (\width, {\hthread*(\thread+0.5)})
            node [at start,black,fill=white,node font=\scriptsize,left,rotate=45,yshift=2mm] {Thread \thread};
    }
    \end{pgfonlayer}
}

\def\drawtimeaxis{
    \draw[arrowtime] (0, \hthread/2) -- (\width, \hthread/2) node
    [midway,fill=white,node font=\small] {Wall Time};
}

\ifx\norenderleft\undefined
\begin{scope}
\pgfmathsetseed{\figseed};
\tikzmath{
\wtot=\wcol+\wimp;
for \thread in {1,...,\nthreads}{
    for \t in {1,...,\ntimesteps}{
        \tm=\t-1;
        \tpct=int(100*\t/\ntimesteps);
        if \thread == 1 then {
            {
                \filldraw[coloredboxsubroutine=\tpct]
                    (\wtot*\tm+\spacing, \hthread+\spacing)
                    rectangle
                    (\wtot*\tm+\wcol-\spacing, {\hthread*(\nthreads+1)-\spacing});
                \node[textsubroutine,rotate=90]
                    (collision\t) at (\wtot*\tm + \wcol/2, {\hthread*(0.5*\nthreads+1)})
                    {Collision Check};
            };
        };
        \x=random(1, 5);
        if \x <= 2 then {{
            \filldraw[coloredboxsubroutine=\tpct]
                (\wtot*\tm+\wcol+\spacing, \hthread*\thread+\spacing)
                rectangle
                (\wtot*\tm+\wtot-\spacing, {\hthread*(\thread+1)-\spacing});
            \node[textsubroutine]
                (impulse\t_\thread) at (\wtot*\tm+\wcol/2+\wtot/2, {\hthread*(\thread+0.5)})
                {Impulse};
        };};
    };
};
\width=\ntimesteps*\wtot;
{\drawtimeaxis};
}
\drawthreadlines
\draw[lineloopbrace]
    (\spacing,{\braceoffset+\hthread*(1+\nthreads)}) -- (\width-\spacing,{\braceoffset+\hthread*(1+\nthreads)})
    node [textloopbrace] {Fused Loop};
\end{scope}
\fi

\tikzmath{
if \vertstack == 0 then {
    {\coordinate (twoloops) at ({1.4+\ntimesteps*(\wcol+\wimp)},0);};
} else {
    {\coordinate (twoloops) at (0,-\hthread*\nthreads-2.5);};
};
}

\ifx\norenderright\undefined
\begin{scope}[shift=(twoloops)]
\pgfmathsetseed{\figseed};
\newcount\ga
\tikzmath{
\maxga=0;
for \thread in {1,...,\nthreads}{
    \ga=0;
    for \t in {1,...,\ntimesteps}{
        \tm=\t-1;
        \wtot=\wcol+\wimp;
        \tpct=int(100*\t/\ntimesteps);
        if \thread == 1 then {
            {
                \filldraw[coloredboxsubroutine=\tpct]
                    (\wcol*\tm+\spacing, \hthread+\spacing)
                    rectangle
                    (\wcol*\tm+\wcol-\spacing, {\hthread*(\nthreads+1)-\spacing});
                \node[textsubroutine,rotate=90]
                    (collision\t) at (\wcol*\tm + \wcol/2, {\hthread*(0.5*\nthreads+1)})
                    {Collision Check};
            };
        };
        \x=random(1, 5);
        if \x <= 2 then {{
            \filldraw[coloredboxsubroutine=\tpct]
                (\ntimesteps*\wcol+\wimp*\ga+\spacing, \hthread*\thread+\spacing)
                rectangle
                (\ntimesteps*\wcol+\wimp*\ga+\wimp-\spacing, {\hthread*(\thread+1)-\spacing});
            \node[textsubroutine]
                (impulse\t_\thread) at (\ntimesteps*\wcol+\wimp*\ga+\wimp/2, {\hthread*(\thread+0.5)})
                {Impulse};
            \global\advance\ga by1
        };};
    };
    \maxga=max(\ga,\maxga);
};
\width=\ntimesteps*\wcol+\maxga*\wimp;
{\drawtimeaxis};
}
\drawthreadlines
\draw[lineloopbrace]
    (\spacing,{\braceoffset+\hthread*(1+\nthreads)}) -- (\ntimesteps*\wcol-\spacing,{\braceoffset+\hthread*(1+\nthreads)})
    node [textloopbrace] {Collision Loop};
\draw[lineloopbrace]
    (\ntimesteps*\wcol+\spacing,{\braceoffset+\hthread*(1+\nthreads)}) -- (\width-\spacing,{\braceoffset+\hthread*(1+\nthreads)})
    node [textloopbrace] {Impulse Loop};
\end{scope}
\fi
\end{tikzpicture}
    \caption{
        Graphical illustration of the execution paths for warp divergence during
        collision detection during a single timestep of simulation. \emph{Left}:
        Execution pattern of a single fused loop, where all threads in a warp
        must wait for even a single impulse to be computed. Performance of this
        approach is shown in \Cref{fig:warp-divergence-performance} as
        \textsc{OneLoop}. \emph{Right}: Execution pattern from precomputing all
        collision checks and storing a list of impulses to be processed and
        then processing them later. Note that this allows impulses from
        different collision checks to be computed simultaneously. Performance of
        this approach is shown in \Cref{fig:warp-divergence-performance} as
        \textsc{TwoLoopsFused} and \textsc{TwoLoopsSplit}.
    }
    \label{fig:warp-divergence-graphic}
\end{figure*}

\subsection{Linked Spatial Hashmaps for Broadphase Nearest Neighbor Searches}
\label{subsec:spatial-hashing}

A naive implementation for detecting collisions between neighboring particles
scales as $O(n^2)$ with the number of particles in the system and would
contribute a significant source of computation time each timestep. To alleviate this, we implement a linked hashmap data structure to generate lists
of candidate collision pairs, which can be built in parallel in \gls{gpu} global
memory. By hashing three-dimensional points according to a discretized spatial
hashing rule, we can quickly check discretized regions of state space for
particle occupancy and, therefore, significantly filter the list of potential
collisions.  We build this data structure atomically in \gls{gpu} memory with a
single \gls{cas} instruction, using \Cref{alg:parallel-hashmap-constructor}.
Since each particle can belong to at most one cell, the entire set of linked
lists can be stored in a single, preallocated array of size $\NParticles$. A
graphical representation of the data structure is represented in
\Cref{fig:spatial-hashmap-graphic}. We hash the discretized coordinates of
individual particles with \Cref{eqn:spatial-hashing}, a modification of the rule
proposed by \citet{teschnerOptimizedSpatialHashing},
\begin{equation}
    \label{eqn:spatial-hashing}
    h(x_i) = \bigoplus_{j=1}^3
    p_j \left(\mathrm{round}\left(\frac{x_{i,j}}{2r}\right) - q\right)
    \quad \left(\mathrm{mod} \; n_h\right),
\end{equation}
where $x_{i,j}$ denotes the $j$th coordinate of the position $x_i \in \R3$, $p_1
    = 73856093$, $p_2 = 19349663$, $p_3 = 83492791$, $q = 100$, $n_h$ is the hashmap
size, $\oplus$ is the \texttt{xor} bitwise logical operator, and $\mathrm{round}$
rounds a real number to the nearest integer.

We find that the size of the hash table has a significant effect on simulator
performance and present results in \Cref{fig:hashtable-benchmark}. As a
heuristic summary of these findings, we propose a hash table size of twice the
number of simulated particles.

We query the hash table in parallel with one thread per particle, traversing the
linked list of particles with the same spatial hash as filtered candidates for
collisions. Since particles can collide across cell boundaries, each thread must
also traverse the $3^3-1=26$ neighboring cell lists. During the linked list
traversal, we check particles for collision against the particle associated with
the current thread. We discuss a method for accelerating narrowphase collision
checking and contact processing on \gls{simt} \glspl{gpu} in
\ref{subsec:warp-divergence}.

\subsection{Collision Geometry using Signed Distance Functions}
\label{subsec:signed-distance-functions}

Our simulation engine models granular materials as particles that are uniform
size and spherical, so narrowphase interparticle collision checking between
particles $i$ and $j$ is as simple as checking $||x_i - x_j||_2^2 \le (2r)^2$.
However, many interesting scenarios for robotic tasks involve objects of
arbitrary geometry, like excavator buckets, bulldozer blades, or robot wheels.
We represent general rigid-body geometry using \glspl{sdf}. A function $f_G :
    \R^3 \to \R$ is a signed distance function for a closed, simple surface $G$ if
$|f_G(x)|$ is the distance between $x$ and the nearest point in $G$. More
formally, given a distance function $d$, $|f_G(x)| = \inf \{d(x, y) \;|\; y \in
    G\}$.  Additionally, the sign of $f$ is positive if $x$ is outside of $G$ and
negative if $x$ is inside of $G$. Using a signed distance function, it is simple
to check collisions with spheres and, thus, to perform particle-body collision
checks. For a sphere of radius $r$ centered at $x$, the maximum penetration
depth into a geometry $G$ is given by \Cref{eqn:sdf-penetration-depth},
\begin{equation}
    \label{eqn:sdf-penetration-depth}
    \psi(x) = f_G(x) - r.
\end{equation}
There are many \glspl{sdf} which can be represented in simple closed forms for
geometric primitives\cite{quilezInigoQuilez}. Of broader interest are geometries
described by arbitrary triangle meshes which are commonly used in physical
simulation.  Computing a \gls{sdf} for an arbitrary mesh can be quite expensive,
depending on the resolution of the mesh surface. To mitigate this, we
pre-compute the \gls{sdf} values on a regular rectilinear three-dimensional grid
which contains the bounds of the mesh geometry. To query the \gls{sdf}, we use a
trilinear interpolation scheme across the knot points in this grid. While this
discretization naturally introduces approximation errors above and beyond the
inherent discretization errors in triangle meshes, such errors are user-tunable
through the grid resolution.  Additionally, while the memory used by the grid
scales $O(n^3)$ with the inverse of the grid resolution, an interpolated lookup
in the grid is always $O(1)$, regardless of the resolution or of the complexity
of the represented geometry. To check for collisions between a world-frame
particle $i$ at coordinates $\InCoordinates{0}{x_i}$ and a rigid body $j$ at
pose $\RigidTransform{0}{j} \in \SEThree$, we transform the particle coordinates
into the local coordinates and evaluate \Cref{eqn:sdf-penetration-depth} as
$\psi\left(\RigidTransform{j}{0}\InCoordinates{0}{x}\right)$, where
$\RigidTransform{j}{0} = \left(\RigidTransform{0}{j}\right)^{-1}$.

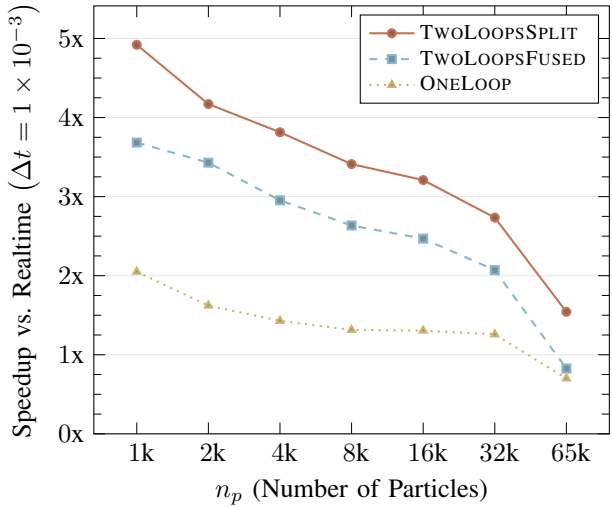
\begin{figure}[t]
    \centering
    \newcommand\plotwarpdivergence[1]{
    \addplot table[
        col sep=comma,
        x={nparticles},
        y={rt_factor_#1},
    ] {figures/data/warpdivergence_archimedes_benchmark_NVIDIA_GeForce_RTX_3080_Ti.csv};
}

\begin{tikzpicture}
\begin{axis}[
    xmode=log,
    speedup plot,
    x axis kilo,
    title={Collision and Contact Algorithm Performance},
    xlabel={$\NParticles$ (Number of Particles)},
    ylabel={Speedup vs. Realtime $\left(\Delta t = \num{1e-3}\right)$},
    ymin={0},
]
    \plotwarpdivergence{twoloops_separate};
    \addlegendentry{\textsc{TwoLoopsSplit}}

    \plotwarpdivergence{twoloops_fused};
    \addlegendentry{\textsc{TwoLoopsFused}}

    \plotwarpdivergence{oneloop};
    \addlegendentry{\textsc{OneLoop}}
\end{axis} 
\end{tikzpicture}
    \caption{
        Speedup vs.\ realtime (higher numbers are better) for different
        orderings for processing contacts and collisions. To maximize the number
        of collision candidates, we simulate particles at rest inside a tall
        cylindrical column.  \textsc{OneLoop} shows a naive implementation,
        while \textsc{TwoLoopsFused} shows the performance of splitting
        collision and contact processing but keeping them in a single kernel.
        \textsc{TwoLoopsSplit} shows the performance of splitting collision and
        contact processing across kernel calls.
    }
    \label{fig:warp-divergence-performance}
\end{figure}

\subsection{Minimizing Warp Divergence}
\label{subsec:warp-divergence}

In \gls{cuda} programs, a \emph{warp} is a collection of threads that execute
the same instruction at the same time. Since each thread sees different data in
the \gls{simt} architecture, conditional instructions may cause two threads in
the same warp to take different branches of control flow, a phenomenon called
\emph{warp divergence}. Since the program counter must remain the same for each
thread in the warp, \gls{cuda} environments will execute both branches of the
conditional, one after the other, and will mask out the effects of instructions
on inactive threads. In effect, warp divergence can substantially negatively
impact program performance on the \gls{gpu}, as threads may spend time ``doing
nothing'' while waiting for other threads to finish their branches.

As suggested in \citet{nakaharaAcceleratingSimulationsGPUs2015}, we minimize
warp divergence by deferring computation of contact impulses until after all
collisions are detected and only marking particle pairs as ``colliding'' during
an initial narrowphase collision detection step. Due to the nature of the
\gls{simt} architecture, splitting the collision checking and contact processing
phases into two loops vastly shortens the number of instructions in the
``colliding'' code path during collision detection.  Despite an efficient
broadphase filter on nearest neighbor search using spatial hashing
(\Cref{subsec:spatial-hashing}), we find experimentally that only about 16\% of
candidate collisions are actually in collision. Since it is significantly faster
to store the indices of colliding pairs than it is to compute the contact
response impulse (\Cref{subsec:projected-jacobi}), we can reduce the time spent
waiting for threads assigned to non-colliding particles. A graphical
representation of the difference in approaches is shown in
\Cref{fig:warp-divergence-graphic}. This performance increase is confirmed
experimentally, using a large pile of particles at rest in a tall cylindrical
column designed to maximize the number of interparticle collisions that are
physically feasible. We show the experimental difference in several approaches
to collision-contact computation order in
\Cref{fig:warp-divergence-performance}.  In summary, we find that splitting the
collision detection and contact impulse computation across two subsequent
\gls{cuda} kernel calls is the fastest approach for all tested numbers of
particles.

\section{Benchmark Environments}
\label{sec:benchmark-environments}
\begin{figure*}[htb]
    \centering
    \input{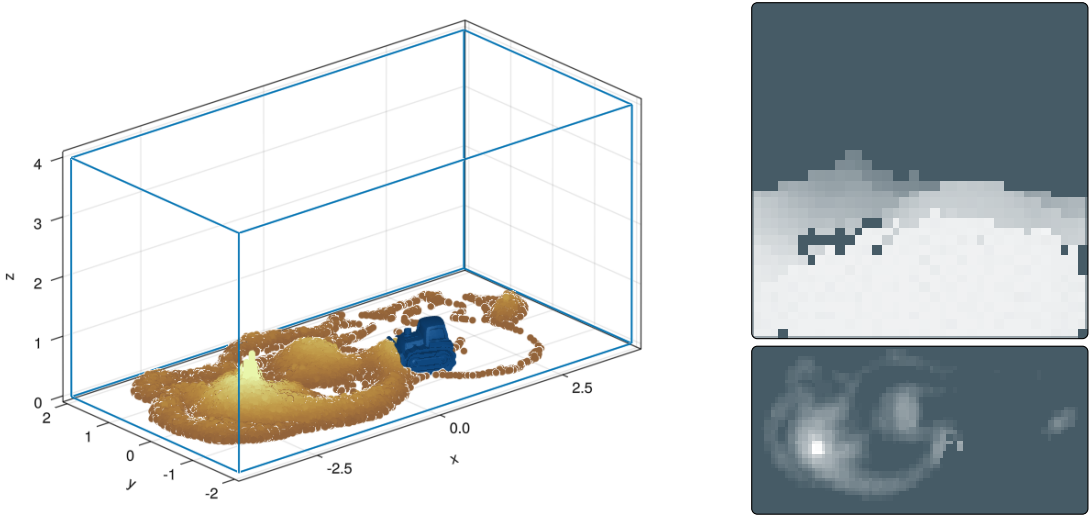}
    \caption{
        On the left is an image taken from the bulldozer simulation environment with 50000
        granular material particles simulated at 1kHz, running in real-time on
        an NVIDIA RTX 3080 Ti \gls{gpu}. Visible here are particles interacting
        with the non-convex geometry of a toy bulldozer model, as well as
        mounding effects from stable interparticle contacts with dry coulomb
        friction. In the top right is displayed a 36-by-36 pixel ``ego view''
        perspective camera showing a depth image from the cockpit of the
        bulldozer, while on the bottom right is a 72-by-36 pixel depth image
        taken from a ``sky view'' orthographic camera. Combined with the lateral
        position and orientation of the bulldozer, these images are used as the
        observation space for a reinforcement learning environment.
    }
    \label{fig:bulldozer-visual}
\end{figure*}

\begin{figure}[htb]
    \centering
    \begin{tikzpicture}

\begin{axis}[
    no markers,
    title={Agent Reward vs.\ Wall Time},
    xlabel={Wall Time (hours)},
    ylabel={Mean Agent Return},
    ymax={7500}
]
    \addplot table[
        col sep=comma,
        x={Relative Wall Time Hours},
        y={Ep Rew Mean},
    ] {figures/data/wandb_export_2023-02-03T02_48_46.378-08_00_reltime.csv};
    \addlegendentry{\acs{td3}}
\end{axis}

\end{tikzpicture}
    \caption{
        A plot of agent total episode return (averaged over the 100 most recent
        episodes) against wall clock time, using the \gls{td3} algorithm. The
        simulation environment, reward structure, and computer hardware used in
        this experiment are described in depth in \cref{subsec:bulldozing}.
    }
    \label{fig:rl-results}
\end{figure}
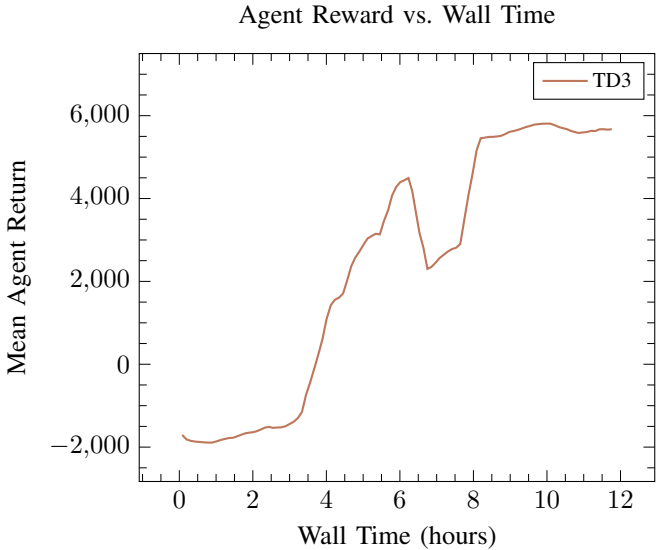

Using the simulation methodology described in \Cref{sec:methods}, we
construct several exemplary simulation environments which showcase various
aspects of our simulation engine, particularly as it pertains to robotics test
and learning environments. We also fully describe the environment we use for
performance benchmarks of our engine.

\subsection{Bulldozing}
\label{subsec:bulldozing}

In the \texttask{Bulldozing} task, shown in \Cref{fig:bulldozer-visual}, the
the controlled robot is a tracked vehicle with a scoop rigidly attached to the
front, modeled after a toy bulldozer STL file. We set up an OpenAI
Gym\cite{brockmanOpenAIGym2016} style environment, designed for a high-achieving
agent to plan a series of actions to use the bulldozer to move material from one
zone to another using a fixed time budget, which we briefly describe in this subsection.

\textpardefn{Dynamics} We use the simulation engine described in this paper, running on
a machine with a 72-core Intel(R) Xeon(R) Gold 6154 CPU and four NVIDIA GeForce
RTX 2080 Ti \glspl{gpu}. We note that our engine currently does not meaningfully
support multiple \glspl{gpu}, besides running multiple independent simulations
simultaneously. On this hardware, our simulation is able to run at realtime with
tens of thousands of particles.  However, to decrease the time required for
training, we sacrifice physical accuracy for computation speed by increasing the
simulation timestep to \SI{0.01}{\second}. We note that this leads to increased
numerical error during integration, leading to nonphysical behavior,
particularly the collapse of tall stacks of particles, and a general behavior of
steep slopes to ``ooze'' down to a more shallow configuration. However, we
believe that the differences in particulate behavior are subtle and that these
errors do not significantly affect the performance of training agents in this
environment. For environments and tasks in which particle stacking and stable
steep slopes are required, this tradeoff may, of course, not be acceptable.

\textpardefn{Observations} We compute two image observations of the environment, both
on the \gls{gpu}. The ``ego camera'' is a perspective depth camera that moves
with the robot mesh and points out of the ``front windshield'' of the
bulldozer.  The ``sky camera'' is an orthographic third-person depth camera
that faces down from the top of the scene and captures the heights of particle
stacks across the entire environment. These images are computed on the same
\gls{gpu} used for simulation, minimizing expensive host-device memory
transfers. Additionally, we provide the $x$ and $y$ positions of the bulldozer,
as well as the yaw angle $\theta$ about the global $z$-axis.

\textpardefn{Actions} We use a track steering model parametrized by the box
$[-1, 1] \times [-1, 1] \subset \R^2$. The first axis describes the linear
velocity of the bulldozer along the direction it is facing, and the second
axis describes the angular velocity around the yaw axis of the bulldozer.

\textpardefn{Rewards} The agent gets a reward of $\frac{100}{n}$, where $n$ is the
total number of simulated particles for each particle inside of a ``goal box''
inside the environment.  To provide a smoother reward function, we assign a
negative value to each particle outside the box of $-\frac{d}{n}$, where $d$ is
the distance from the particle to the closest point on the box.

We train this environment using the \gls{td3} algorithm and show that an agent
is able to learn to achieve higher rewards than an initial random policy. We do
not present these results as an exemplary demonstration of robot learning, but
instead to show that complex environments can be simulated fast enough to train
Reinforcement Learning agents in days.

\subsection{Helical Gear Tower}
\label{subsec:helical-gear-tower}

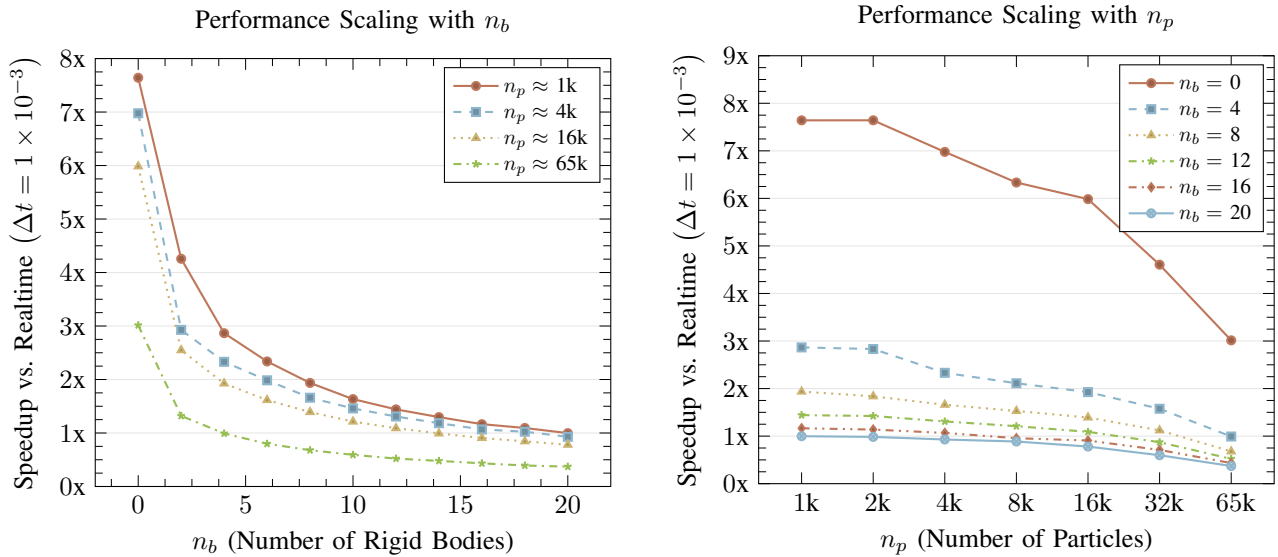
\begin{figure*}[htb]
    \centering
    \newcommand\plotmatching[3]{
    \addplot table[
        col sep=comma,
        x={#1},
        y=rt_factor,
        discard if not={#2}{#3},
    ] {figures/data/archimedes_benchmark_NVIDIA_GeForce_RTX_3080_Ti.csv};
}

\begin{tikzpicture}
\begin{axis}[
    speedup plot,
    title={Performance Scaling with $\NBodies$},
    xlabel={$\NBodies$ (Number of Rigid Bodies)},
    ylabel={Speedup vs. Realtime $\left(\Delta t = \num{1e-3}\right)$},
    ymin={0},
    ymax={8},
]
    \foreach \nparticles in {1024,4096,16384,65536} {
        \pgfkeys{/pgf/fpu=true,/pgf/fpu/output format=fixed}
        \tikzmath{ \npthoufp = int(\nparticles/1000); }
        \pgfkeys{/pgf/fpu=false}
        \tikzmath{ \npthou = int(\npthoufp); }
        
        \edef\dolegend{\noexpand\addlegendentry{$\NParticles \approx$ \npthou k}} 
        \plotmatching{nscrews}{nparticles}{\nparticles};
        \dolegend
    }
\end{axis} 
\end{tikzpicture}%
\qquad%
\begin{tikzpicture}
\begin{semilogxaxis}[
    speedup plot,
    x axis kilo,
    title={Performance Scaling with $\NParticles$},
    xlabel={$\NParticles$ (Number of Particles)},
    ylabel={Speedup vs. Realtime $\left(\Delta t = \num{1e-3}\right)$},
    ymin={0},
    ymax={9},
]
    \foreach \nbodies in {0,4,8,12,16,20} {
        \edef\dolegend{\noexpand\addlegendentry{$\NBodies =$ \nbodies}} 
        \plotmatching{nparticles}{nscrews}{\nbodies};
        \dolegend
    }
\end{semilogxaxis} 
\end{tikzpicture}
    \caption{
        Plots of speedup vs.\ realtime (higher numbers are better) against the
        number of rigid bodies in the scene (left) and the number of particles
        simulated (right, note log $x$ axis).  The benchmark environment
        simulated is described in \Cref{subsec:helical-gear-tower}.
    }
    \label{fig:helical-gear-benchmark}
\end{figure*}

\begin{figure}[htb]
    \centering
    \begin{tikzpicture}
        \node [inner sep=0pt] at (centerpoint) {\includegraphics[height=8.0cm]{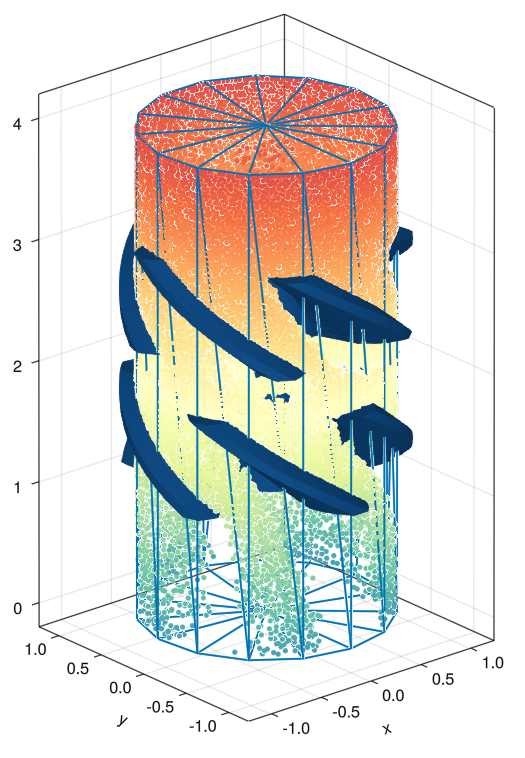}};
    \end{tikzpicture}
    \caption{
        Image of the helical gear benchmark setup that we use to test
        performance as it scales with the number of bodies and the number of particles.
        Particles are colored according to their $z$-axis coordinate.
    }
    \label{fig:helical-gear-image}
\end{figure}

To test our performance and demonstrate our ability to scale to multiple rigid
bodies of arbitrary, non-convex geometry, we simulate a tall, cylindrical
containment tower with a varying number of involute helical gears rotating at
uniformly random velocities, as shown in \Cref{fig:helical-gear-image}. To keep
particles flowing, we implement a cyclic boundary condition at the vertical
extents of the cylinder, where particles passing through the bottom cap are
transported to the top of the cylinder with the same velocity. However, we do
not implement this cyclic boundary for nearest neighbor contact searches. We
present results of how performance scales with the number of particles and the number of
rigid bodies in \Cref{fig:helical-gear-benchmark}. While we do not propose a
quantitative measurement of the ``complexity'' of a mesh geometry, we select
these helical gears as a benchmark of interaction with arbitrary non-convex
geometry, and note that due to our gridded \gls{sdf} representation of collision geometries discussed in \cref{subsec:signed-distance-functions}, our performance
does not change with the shape or complexity of the mesh.

\subsection{Excavation}
\label{subsec:excavation}

In the \texttask{Excavation} environment, shown in \Cref{fig:hero-image}, the
robotic manipulator is a 7-\gls{dof} Franka Emika Panda robot with a tool
rigidly attached that is similar in shape to an excavator bucket. The robot
accepts velocity control inputs in joint space. In our testing, we are able to
interact with 50000 particles at realtime speeds on a single NVIDIA GeForce RTX
3080 Ti, using a simulation timestep of \num{5e-4}.

We select the Franka Emika Panda robot for its flexibility and because it is
becoming commonplace in the robotic research world. However, the Panda
highlights an important limitation of our simulation, the fact that rigid bodies
are kinematically driven and do not accumulate reaction forces from the
environment. Particularly in granular manipulation tasks, there is a stark difference in the amount of force required to push a cutting edge through an
aggregate (the \emph{penetration force}), and the amount of force required to
push a flat surface, like the back of a shovel, through an aggregate (the
\emph{deadload force}). Since the Panda robot is designed for safety around
humans and in research applications, it has a relatively low maximum applicable
force, leading to a wide gap in behavior between simulation and reality, which
we intend to address in future improvements to our engine. Currently, users may
manually compute the total force on an individual particle or rigid body, and use this in a higher-level decision to terminate an action or training episode,
which we propose as a mitigation for this limitation.

\section{Related Works}
\label{sec:related-works}
Simulation of granular dynamics is of great interest for many tasks besides
robotics, including Computer Aided Design and Engineering of earthmoving
equipment~\cite{nezamiSimulationFrontEnd2007}, scientific sampling apparatus
~\cite{riccobonoGranularFlowCharacterization2021}, and production lines
~\cite{geerMoreEfficientMethod2018}. Accordingly, there are several overarching
approaches for the simulation of granular dynamics in existing literature, each
with various design decisions that affect computational efficiency and accuracy.
Broadly, these are \glsfirst{dem}, which represents a Lagrangian state of granular
materials as a set of distinct bodies, Computational Fluid Dynamics, or
\gls{mpm} approaches, which use a hybrid Eulerian-Lagrangian state
representation, and Reactive Force Theories, which are extremely efficient and
use a fully data-driven approach, but only model the force on a tool exerted
by soil. Our presented framework falls firmly into the \acrfull{dem} category.

The simulation of bulk material by the interactions of discrete particles with
Lagrangian state is generally called the \acrfull{dem}.  One popular \gls{dem}
framework for granular dynamics is \gls{lammps}
~\cite{thompsonLAMMPSFlexibleSimulation2022}, which offers a ``Granular'' package
for rigid or elastic interparticle physical contact forces.  A fork of
\gls{lammps}, called \gls{liggghts}~\cite{klossModelsAlgorithmsValidation2012},
also offers large-scale granular dynamics simulations with arbitrary particle
geometries and sizes, as well as interactions with arbitrarily shaped rigid
objects. While both \gls{lammps} and \gls{liggghts} are well-tested and proven
simulation codes that offer significant flexibility to model multiple physical
models of contact, they have been primarily designed for large clusters of CPUs
communicating with networked message-passing interfaces, and are inherently
limited in their ability to reach ``interactive'' rates of simulation for large
numbers of particles.

Other solvers, such as Ansys Rocky~\cite{AnsysRockyParticle}, target granular
interactions on \glspl{gpu}, and are flexible, high quality, and high
performance. However, these systems are closed-source simulation codes without
published methods, which require expensive licenses, and prohibit user
modification of the underlying software. Algoryx Momentum
Granular~\cite{servinMultiscaleModelTerrain2021a} provides real-time simulation
of granular materials, along with a two-way coupling with rigid multibody
physics, by using an adaptive multiscale particle model, presumably
executing on a CPU. While the authors publish their methodology, the
implementation is also closed source and available only under a commercial
license.

Project Chrono is a multiphysics engine supporting multibody rigid dynamics,
computational fluid dynamics, and solid mechanics through finite element
analysis. Project Chrono also offers
Chrono::GPU~\cite{fangChronoGPUOpenSource2021}, a \gls{gpu} accelerated engine
for granular flows. Chrono::GPU does not target realtime interaction, instead
supporting elastic contact modes for very high accuracy simulations, which run
significantly slower than realtime.

Wheel-soil interaction, also called terramechanics, is a key aspect
of modeling wheeled robots traversing off-road terrains. This behavior is
especially important during highly dynamic motions or when traversing loose or
rocky soils. Previous approaches have found success by directly learning
input-output maps of vehicle dynamics on such terrains from real data from
trajectories from human experts and then used the learned model directly for
Model-Predictive Path Integral control schemes to achieve ``drifting'' like
behavior on dirt racetracks~\cite{williamsAggressiveDrivingModel2016}.

Additionally, there have been several efforts in robotics to control
heavy earth-moving equipment like excavators. Early efforts used a physically
based analytical model of soil computing onlly reactive forces of the earth on
the bucket mechanism of the excavator~\cite{parkDevelopmentVirtualReality2002}.
Similarly, more recent efforts have achieved impressive results by using
reinforcement learning to control excavator buckets in inhomogeneous soil
densities by using a reduced analytical reactive force model, and employing
heavy domain randomization to train a reinforcement learning policy for bucket
control~\cite{egliSoilAdaptiveExcavationUsing2022}. A key stated intention for
the use of reduced, force-based models in these works is the computational
intensity required for a full, Lagrangian-state particle simulation of soil
mechanics. Reactive force models are limited in their ability to richly
represent the state of the soil under manipulation, and are thus inapplicable to
completely simulate tasks such as material transport.

Other approaches include the \gls{mpm}, which models Lagrangian particle states
but integrates their dynamics on Eulerian grids. \gls{mpm} is an extremely flexible modeling approach for a wide variety of physical phenomena, but in
general, is not efficient to run at interactive simulation
speeds\cite{haeriEfficientNumericalMethods2020}, despite high quality
implementations available as part of the
DiffTaichi\cite{huDifftaichiDifferentiableProgramming2019} GPU framework.

There are many high quality interfaces for GPU programming in high level
languages,  including the aforementioned Taichi framework and the
Warp\cite{macklinWarpHighperformancePython2022} library from NVIDIA. We choose
the Julia language due to its composable interfaces, high quality GPU
libraries\cite{besard2018juliagpu}, and the native compilation of non-GPU code.

\section{Conclusion and Future Work}
\label{sec:conclusion}
We present a methodology for a simulation framework of hundreds of thousands of
particles in dry frictional contact, which in bulk approximate granular media.
Our framework is for execution on \gls{cuda} \glspl{gpu}, and we describe
methods for high performance on such architectures.  While our approach does not
model all types of materials, our engine favors speed while modeling a
Lagrangian state, which is critical for robotic tasks that need to manipulate
the state of soil. A performant implementation of the simulation architecture,
which was used for the performance benchmarks in this paper is written in the
Julia programming language and is released as open-source software\footnote{
    Link withheld for review anonymity.
}. This framework is able to handle one-way contact coupling with rheonomically
constrained rigid bodies of arbitrary geometry, and allows a user-tunable
tradeoff between geometry accuracy and \gls{cuda} memory usage, while
maintaining constant compute time performance for rigid body collision
detection.

Our model of granular interaction incorporates several simplifications and is
not an appropriate model for all of the various types of granular interactions
present in nature. Particularly in robotics, however, where fast simulation
allows planning algorithms to predict dynamics in closed-loop control
algorithms and enables fast environments for robot learning setups, we believe
there is a role for a granular simulation engine that is focused on speed and
which can represent arbitrary state configurations of particulate media.

To demonstrate the utility of our simulation framework for tasks in robotics,
we show a collection of several benchmark robotic tasks involving
the manipulation of granular materials, as well as showing that state of the art reinforcement learning algorithms can be brought to bear against these problems. An open-source implementation of these environments, along with an OpenAI Gym-like API, is available in the linked software repository.

We hope that these contributions spur increased research in the difficult research area of robotic manipulation of granular materials, and see several
clear paths toward a more complete simulation framework, which we list in
the following paragraphs and intend to address in future work.

\textpardefn{Two-Way Coupling} For tasks where the mass of the accumulated material
is non-negligible compared to the actuation power of the manipulator, a full
rigid-body or multiphysics dynamics engine, and a full two-way force
coupling is required.

\textpardefn{Differentiability} Many robotic dynamic planning algorithms
and system identification algorithms use Jacobians of the dynamics to
accelerate optimization. The Lagrangian particle state representation is not
amenable to useful direct differentiation, due to its permutation
invariance. A carefully chosen state representation could pave a path
toward meaningful differentiability.

\textpardefn{Non-Homogeneity} While the presented framework allows for an
easily replaceable interparticle contact model, each particle has the same
parameters and contact equations. Since many real-world materials are
non-homogenous, efficient computation for such media is an important
contribution towards a fully flexible simulator.

\section*{Acknowledgments}
\label{sec:acknowledgements}
This work is supported by the NASA Space Technology Research Fellowship, grant
number 80NSSC19K1182.
The authors thank Lorenzo Fluckiger, Trey Smith, Brian Coltin, Gautam Salhotra,
K.R.\ Zenter, and Shashank Hegde for helpful discussions and feedback during the
writing of this paper.

\bibliographystyle{plainnat}
\bibliography{references}

\ifoptionfinal{}{
    \listoftodos
}

\end{document}